\documentclass{article}
\usepackage{paperstyle,times}
\usepackage{amsmath,amssymb,amsthm}
\usepackage{graphicx,booktabs,array,multirow}
\usepackage{microtype}
\usepackage{subcaption}
\usepackage{float}
\usepackage{xcolor}
\usepackage{tikz}
\usetikzlibrary{positioning,arrows.meta,calc,decorations.pathreplacing}
\usepackage{hyperref}
\usepackage{url}
\hypersetup{hidelinks}

\finalcopy

\newcommand{\blfootnote}[1]{%
  \begingroup
  \renewcommand{\thefootnote}{}\footnote{#1}%
  \addtocounter{footnote}{-1}%
  \endgroup
}

\newtheorem{theorem}{Theorem}
\newtheorem{proposition}{Proposition}
\newtheorem{corollary}{Corollary}
\newtheorem{assumption}{Assumption}
\newtheorem{hypothesis}{Hypothesis}

\definecolor{slres}{HTML}{E8F3EC}
\definecolor{slresb}{HTML}{2E7D4F}
\definecolor{slinv}{HTML}{FDF0E1}
\definecolor{slinvb}{HTML}{C0661A}
\definecolor{sldiag}{HTML}{E6F0F8}
\definecolor{sldiagb}{HTML}{1F5F95}
\definecolor{slgrey}{HTML}{6B6B6B}

\title{Neural Succession:\\
A Mesoscopic Theory of Invasion, Coexistence, and Stabilization in Continual Learning}
\author{
\hspace*{-\tabcolsep}Shoaib Ahmed Dipu$^{1}$, Md Salman Shamil$^{2}$, Sayeed Shafayet Chowdhury$^{1}$ \\[4pt]
\hspace*{-\tabcolsep}{\normalfont\normalsize $^{1}$Indiana University Indianapolis \quad $^{2}$North South University} \\[2pt]
\hspace*{-\tabcolsep}{\normalfont\small\texttt{\{shdipu,\,saychow\}@iu.edu} \quad \texttt{salman.shamil@northsouth.edu}}
}

\begin{document}
\maketitle
\lhead{}\chead{}\rhead{}
\renewcommand{\headrulewidth}{0pt}
\blfootnote{Code is available at \url{https://github.com/shoaibdipu/Neural_Succession/}.}

\begin{abstract}
Continual learning is usually studied through mechanisms that preserve old knowledge. Here we study what happens when a new task enters a representation that already supports earlier tasks. We develop \emph{Successional Learning Theory} (SLT), a mesoscopic account in which the current representation is a resident community, the incoming task is an invader, forgetting is resident displacement, joint retention is coexistence, replay is resident reinforcement, and training moves from establishment toward stabilization. Its empirical coordinate is directional pre-invasion compatibility, measured on the resident model before the incoming task is learned. Across eight experiments, compatibility orders later forgetting on the 20 directed Split-CIFAR-10 transitions (three-repeat $r=-0.789$, incoming-task cluster 95\% CI $[-0.90,-0.72]$; every repeat alone $r\le-0.67$), forecasts held-out forgetting with 24\% lower error than a no-information baseline, and reproduces under controlled MNIST permutations and CIFAR-10 rotations ($r=-0.804$, $-0.718$). The same coordinate also resolves coexistence. The three natural transitions that coexist are exactly the three most compatible (AUC $1.00$). On an 84-transition suite, compatibility separates coexistence from exclusion at every retention threshold (AUC $0.93$--$0.97$). Replay repairs every transition with at most 325 stored examples and is most efficient where displacement is largest. The predictive structure appears at the transition scale. Compatibility reaches $|r|=0.720$, while activation, representation, Jacobian, and fixed-coefficient Lotka--Volterra specializations do not. Plasticity and feature turnover fall reliably from early to late training (15/15 and 14/15 runs). We formalize a minimum habitat-modification bound, a displacement floor, a sufficient coexistence condition, an identifiability law with a range-restriction corollary, successional stabilization, and local reinforcement. The identifiability law also predicts where the coordinate loses leverage, and the prediction matches three CIFAR-100 partitions and five optimizer regimes. SLT is a pre-adaptation diagnostic that complements replay, regularization, and projection methods.
\end{abstract}

\section{Introduction}
Continual learning modifies a model that already supports earlier tasks \citep{mccloskey1989catastrophic,french1999catastrophic,goodfellow2013empirical}. When task $B$ arrives after task $A$, it is not learned in an empty parameter space. It enters a representation shaped by the resident task. Some incoming tasks reuse what is already present and cause little damage. Others require broader reorganization and displace previous behavior. Most continual-learning work starts after this transition has begun. It asks how to regularize parameters \citep{kirkpatrick2017overcoming,zenke2017si,aljundi2018mas,li2017lwf}, constrain or project gradients \citep{lopezpaz2017gem,chaudhry2019agem,farajtabar2020ogd,saha2021gpm,zeng2019owm,wang2021adamnscl}, replay old examples \citep{rebuffi2017icarl,rolnick2019experience,chaudhry2019tiny,buzzega2020dark}, or isolate capacity \citep{rusu2016progressive,mallya2018packnet}; see \citet{parisi2019review,delange2022survey,wang2024survey} for surveys. We instead study the \emph{transition itself}.

We call this view \emph{Successional Learning Theory} (SLT). It borrows its questions from ecological succession, the ordered replacement of communities in which early occupants modify the habitat that later arrivals must use \citep{clements1916succession,odum1969strategy,connell1977succession,odlingsmee2013niche}. In this view, a learned representation is the resident habitat and an incoming task is an invader. The representation change required to solve it is establishment demand. Damage to resident-sensitive directions is displacement, joint retention is coexistence, and replay reintroduces resident signal. Within-task dynamics move from high-plasticity establishment toward stabilization. The ecology supplies a linked set of questions. Every formal object is defined directly through neural risk, representation change, or optimization.

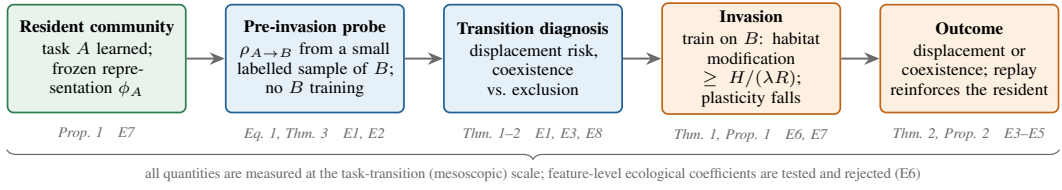
\begin{figure}[t]
\centering
\resizebox{\linewidth}{!}{%
\begin{tikzpicture}[
  box/.style={rectangle, rounded corners=2.5pt, draw=#1, line width=0.7pt, align=center, minimum height=1.55cm, text width=2.35cm, inner sep=3pt, font=\scriptsize},
  arr/.style={-{Stealth[length=2.2mm,width=1.8mm]}, line width=0.8pt, color=slgrey},
  lab/.style={font=\tiny\itshape, text=slgrey, align=center, text width=2.4cm}]
\node[box=slresb, fill=slres] (res) {\textbf{Resident community}\\[1pt] task $A$ learned;\\ frozen representation $\phi_A$};
\node[box=sldiagb, fill=sldiag, right=0.55cm of res] (probe) {\textbf{Pre-invasion probe}\\[1pt] $\rho_{A\to B}$ from a small\\ labelled sample of $B$;\\ no $B$ training};
\node[box=sldiagb, fill=sldiag, right=0.55cm of probe] (diag) {\textbf{Transition diagnosis}\\[1pt] displacement risk,\\ coexistence vs.\ exclusion};
\node[box=slinvb, fill=slinv, right=0.55cm of diag] (inv) {\textbf{Invasion}\\[1pt] train on $B$: habitat\\ modification $\geq H/(\lambda R)$;\\ plasticity falls};
\node[box=slinvb, fill=slinv, right=0.55cm of inv] (out) {\textbf{Outcome}\\[1pt] displacement or\\ coexistence; replay\\ reinforces the resident};
\draw[arr] (res) -- (probe); \draw[arr] (probe) -- (diag); \draw[arr] (diag) -- (inv); \draw[arr] (inv) -- (out);
\node[lab, below=0.08cm of res] {Prop.~\ref{prop:stabilization}\quad E7};
\node[lab, below=0.08cm of probe] {Eq.~\ref{eq:rho}, Thm.~\ref{thm:ident}\quad E1, E2};
\node[lab, below=0.08cm of diag] {Thm.~\ref{thm:habitat}--\ref{thm:coexist}\quad E1, E3, E8};
\node[lab, below=0.08cm of inv] {Thm.~\ref{thm:habitat}, Prop.~\ref{prop:stabilization}\quad E6, E7};
\node[lab, below=0.08cm of out] {Thm.~\ref{thm:coexist}, Prop.~\ref{prop:reinforce}\quad E3--E5};
\draw[decorate, decoration={brace, mirror, amplitude=4pt}, color=slgrey, line width=0.5pt]
  ($(res.south west)+(0,-0.52)$) -- ($(out.south east)+(0,-0.52)$)
  node[midway, below=4pt, font=\tiny, text=slgrey] {all quantities are measured at the task-transition (mesoscopic) scale; feature-level ecological coefficients are tested and rejected (E6)};
\end{tikzpicture}}
\caption{\textbf{The SLT pipeline.} A resident representation is probed \emph{before} the incoming task is learned. The resulting coordinate diagnoses the transition, which then unfolds as establishment followed by stabilization and ends in displacement or coexistence. Grey labels map each stage to the result that formalizes it and the experiment that tests it.}
\label{fig:framework}
\end{figure}

The distinction from task-similarity work is scope. Task relation is known to matter for transfer, forgetting, and task ordering \citep{nguyen2019toward,ramasesh2021anatomy,lee2021teacher,doan2021ntk,hiratani2024disentangling,li2025optimal,wang2025spot,wang2026gnr}. SLT does not claim that a transfer score is new. It asks whether one pre-invasion coordinate can organize outcomes that are usually analyzed separately, including resident displacement, coexistence versus exclusion, response to resident reinforcement, and the move from establishment to stabilization. We also test the neural scale at which this ecological interpretation is predictive.

Across these tests, pre-invasion compatibility orders forgetting on natural transitions and under controlled manipulations. It also forecasts held-out forgetting and separates coexistence from exclusion. It is more predictive than activation covariance, representation similarity, raw Jacobian overlap, and stationary microscopic ecological specializations. Within an invasion, plasticity and feature turnover fall from early to late training. The theory predicts where a prospective coordinate should lose leverage, and the data follow this prediction. Our contributions are:
\begin{itemize}
\item \textbf{A mesoscopic theory of neural succession} (\S\ref{sec:theory}): a minimum habitat-modification bound, a displacement floor, a sufficient coexistence condition, an identifiability law with a range-restriction corollary, successional stabilization, and a local reinforcement result.
\item \textbf{A prospective compatibility principle} (\S\ref{sec:e1}): a directional score measured before learning $B$ orders later forgetting, supports held-out prediction, and reproduces under controlled task manipulations.
\item \textbf{Phase, scale, and stage results} (\S\ref{sec:e3}--\ref{sec:e7}): the coordinate separates coexistence from exclusion, dominates the tested microscopic specializations, and adaptation shows a reproducible establishment-to-stabilization profile.
\item \textbf{Robustness and a predictive scope law} (\S\ref{sec:scope}): the ordering holds across estimator settings, label budgets, optimizer regimes, and a pretrained backbone, and the identifiability law predicts the regimes in which it loses leverage.
\end{itemize}

\section{Related work and positioning}
\label{sec:related}
\paragraph{Invasion and displacement.} Theory links forgetting to task geometry through NTK overlap \citep{jacot2018ntk,doan2021ntk}, teacher--student similarity \citep{lee2021teacher,hiratani2024disentangling}, linear-regression analyses \citep{evron2022catastrophic,lin2023theory}, and representation-level anatomy \citep{ramasesh2021anatomy}. Computing optimal continual learners is intractable in general \citep{knoblauch2020optimal}. Transferability scores estimate how well frozen features support a new task \citep{tran2019nce,nguyen2020leep,you2021logme}, and fine-tuning can distort useful features \citep{kumar2022finetuning}. Task relations have also been used to predict forgetting risk \citep{nguyen2019toward,wang2025spot}, to regularize dissimilar tasks \citep{wang2026gnr}, to route similar and dissimilar tasks \citep{ke2020mixed,wang2022sustainable,adel2024similarity}, and to order tasks \citep{bell2022ordering,li2025optimal}. Our score follows the normalized-transfer construction of \citet{li2025optimal}. SLT adds a resident-side consequence, a lower bound on the habitat modification that establishment demands (Theorem~\ref{thm:habitat}).
\paragraph{Coexistence.} In ecology, coexistence requires sufficient niche differentiation \citep{gause1934struggle,hardin1960competitive,macarthur1967limiting,tilman1982resource,chesson2000mechanisms}. \citet{meszena2006limiting} separate a species' impact on regulating factors from its sensitivity to them. The ML analogue is protected-subspace learning, which steers new updates away from directions that matter for old tasks \citep{zeng2019owm,farajtabar2020ogd,saha2021gpm,wang2021adamnscl}. Theorem~\ref{thm:coexist} turns this shared intuition into a sufficient condition phrased in resident-sensitive displacement.
\paragraph{Reinforcement.} Invasion biology identifies propagule pressure as a main determinant of establishment \citep{lockwood2005propagule}, and resident communities resist invasion \citep{levine2004biotic}. In continual learning, replay and gradient constraints protect the resident \citep{lopezpaz2017gem,chaudhry2019agem,riemer2019mer,aljundi2019mir,buzzega2020dark,seo2025budgeted}. Proposition~\ref{prop:reinforce} states the local first-order condition under which reintroduced resident signal counters displacement.
\paragraph{Successional stages.} Deep networks change quickly early in training and stabilize later \citep{achille2019critical,frankle2020early,gurari2018tiny}, and continual training erodes plasticity \citep{ash2020warm,lyle2023plasticity,dohare2024plasticity}. Training regime modulates forgetting \citep{mirzadeh2020regimes,mirzadeh2021lmc}. We measure feature-level plasticity and turnover within a transition and give a sufficient condition for finite total feature variation.
\paragraph{Ecology as a model of learning.} Some learning systems admit exact ecological dynamics \citep{howell2020ecology}, and Lotka--Volterra competition \citep{lotka1925elements,volterra1926fluctuations} is a natural candidate for feature interaction. We test that candidate directly (E6) and find that the predictive ecological structure lives at the transition scale, consistent with ecological evidence that coarse community-level variables can be more predictive than pairwise coefficients \citep{mayfield2010opposing}.

\begin{table}[t]
\caption{Positioning by the question each family answers. ``Pre'': computable before task-$B$ adaptation.}
\label{tab:position}
\centering
\scriptsize
\setlength{\tabcolsep}{2.4pt}
\begin{tabular}{p{1.18in}p{1.18in}cccc}
\toprule
Family & Main object & Pre & Coexistence & Scale test & Stages \\
\midrule
Transferability / SPOT / NTK & task relation, kernel & $\checkmark$ & & & \\
EWC / SI / MAS & parameter importance & & & & \\
GEM / OGD / GPM & gradient subspaces & & implicit & & \\
Replay / DER++ & stored examples, logits & & & & \\
Task ordering & sequence relation & $\checkmark$ & & & \\
Plasticity studies & training dynamics & & & & $\checkmark$ \\
\textbf{SLT (ours)} & \textbf{resident--invader transition} & $\checkmark$ & $\checkmark$ & $\checkmark$ & $\checkmark$ \\
\bottomrule
\end{tabular}
\end{table}

\section{Successional Learning Theory}
\label{sec:theory}
\subsection{Pre-invasion compatibility and establishment}
Following \citet{li2025optimal}, the directional pre-invasion score is
\begin{equation}
\rho_{A\to B}=1-\sqrt{\frac{e_B(\theta_A)}{e_{B,\mathrm{sf}}(\theta_A)}},
\qquad \Omega_{A\to B}=1-\rho_{A\to B},
\label{eq:rho}
\end{equation}
where $e_B(\theta_A)$ is the error of a constrained incoming-task probe on the frozen resident representation and $e_{B,\mathrm{sf}}(\theta_A)$ is a shuffled-label reference. Larger $\rho$ means $B$ is already better supported. We treat $\Omega$ as an empirical coordinate and do \emph{not} identify it with the theoretical demand below.

\begin{hypothesis}[Successional compatibility principle]
\label{hyp:scp}
Within a fixed task ecology and training regime, expected resident forgetting $\mathbb E[F_{A\to B}\mid\Omega_{A\to B}]$ is increasing in $\Omega_{A\to B}$.
\end{hypothesis}

Let $\phi_A$ be the resident representation and
\begin{equation}
\mathcal R_B^{(R_B)}(\phi)=\inf_{\|w\|\le R_B}\mathbb E_B\,\ell(w^\top\phi(x),y),\qquad
H_{A\to B}=\big[\mathcal R_B^{(R_B)}(\phi_A)-\ell_B^\star\big]_+ .
\end{equation}
Let $\mathcal E_B$ be the reachable representations whose constrained probe risk is at most $\ell_B^\star$, and $D_B(\phi)=\big(\mathbb E_B\|\phi(x)-\phi_A(x)\|^2\big)^{1/2}$.

\begin{theorem}[Minimum habitat modification]
\label{thm:habitat}
If $\ell$ is $\lambda_B$-Lipschitz in the logits, every $\phi\in\mathcal E_B$ satisfies $D_B(\phi)\ge H_{A\to B}/(\lambda_B R_B)$.
\end{theorem}
An unsupported invader must change the representation before it can establish. The required change grows with its establishment deficit. Let $S_A(\phi)$ measure resident-sensitive displacement.
\begin{assumption}[Resident sensitivity band]
\label{ass:sens}
On the reachable region there are $0<m_A\le M_A<\infty$ with
$\frac{m_A}{2}S_A(\phi)^2\le [L_A(\phi)-L_A(\phi_A)]_+\le \frac{M_A}{2}S_A(\phi)^2$.
\end{assumption}
Define $\sigma_{A\to B}=\inf_{\phi\in\mathcal E_B,D_B(\phi)>0}S_A(\phi)/D_B(\phi)$ and $\Pi_{A\to B}=\sigma_{A\to B}H_{A\to B}/(\lambda_BR_B)$.
\begin{corollary}[Displacement floor]
\label{cor:floor}
Every establishing representation satisfies $S_A(\phi)\ge\Pi_{A\to B}$, hence $[L_A(\phi)-L_A(\phi_A)]_+\ge \frac{m_A}{2}\Pi_{A\to B}^2$. If $\Pi_{A\to B}>\sqrt{2\delta_A/m_A}$, no $B$-establishing representation keeps $A$ within loss tolerance $\delta_A$.
\end{corollary}
\emph{Testable consequence.} The floor increases with $H_{A\to B}$; if $\Omega$ tracks $H$, forgetting should increase with $\Omega$ (Hypothesis~\ref{hyp:scp}). E1, E2, and E8 test the ordering, and a bridge analysis tests whether $\Omega$ tracks $H$.

\subsection{Coexistence}
Let $s_{A\to B}^\star=\inf_{\phi\in\mathcal E_B}S_A(\phi)$ be the least resident-sensitive displacement compatible with establishment.
\begin{theorem}[Niche-separable coexistence]
\label{thm:coexist}
Under Assumption~\ref{ass:sens}, if $s_{A\to B}^\star\le\sqrt{2\delta_A/M_A}$, then for every $\epsilon>0$ some reachable representation establishes $B$ while increasing resident loss by at most $\delta_A+\epsilon$ (at most $\delta_A$ if the infimum is attained).
\end{theorem}
Corollary~\ref{cor:floor} and Theorem~\ref{thm:coexist} define 2 sides of the phase picture. Small resident-sensitive establishment cost permits coexistence. Large unavoidable displacement forces exclusion. $\sigma$ and $s^\star$ are latent, so the prediction is tested behaviorally (E3).

\subsection{When a prospective coordinate can work}
\begin{theorem}[Identifiability]
\label{thm:ident}
Within a fixed ecology, let $F=\alpha+\beta\Omega+\varepsilon$ with $\mathbb E[\varepsilon\mid\Omega]=0$ and $\operatorname{Var}(\varepsilon)=\sigma_\varepsilon^2$. Then $R^2(F\mid\Omega)=\beta^2\sigma_\Omega^2/(\beta^2\sigma_\Omega^2+\sigma_\varepsilon^2)$. If $\widetilde\Omega=\Omega+\eta$ with independent noise of variance $\sigma_\eta^2$, the squared correlation is multiplied by $\sigma_\Omega^2/(\sigma_\Omega^2+\sigma_\eta^2)$.
\end{theorem}
\begin{corollary}[Range restriction]
\label{cor:range}
Suppose two ecologies share $\beta$ and $\sigma_\varepsilon$ but differ in the spread of the coordinate, $\sigma'_\Omega=s\,\sigma_\Omega$. If $r$ is the correlation in the first, the correlation in the second is
$|r'|=|r|\,s/\sqrt{1-r^2+r^2s^2}$.
\end{corollary}
Corollary~\ref{cor:range} is the classical range-restriction relation \citep{thorndike1949personnel}. It gives a quantitative prediction for when the score should weaken. If transitions in an ecology are nearly equally compatible, the ordering must weaken even when the underlying mechanism is unchanged (\S\ref{sec:scope}).

\subsection{Stabilization and reinforcement}
\begin{proposition}[Finite feature variation under convergent training]
\label{prop:stabilization}
Suppose single-task training follows gradient flow on an $L_s$-smooth objective satisfying a $\mu$-Polyak--{\L}ojasiewicz inequality \citep{karimi2016pl}, and active feature statistics $x_i(\theta)>0$ satisfy $\|\nabla_\theta\log x_i(\theta)\|\le C_i$. Then $\int_0^\infty|\tfrac{d}{dt}\log x_i(\theta_t)|\,dt<\infty$.
\end{proposition}
Total feature change is finite, so most change must occur early. Establishment precedes stabilization (E7). This is a sufficient idealized result, not a claim that ResNet training is globally PL.

\begin{proposition}[Local resident reinforcement]
\label{prop:reinforce}
Consider one gradient step of size $\eta$ on $L_q=(1-q)L_B+qL_A$, $q\in[0,1]$, with $g_A=\nabla L_A$, $g_B=\nabla L_B$, $a=\|g_A\|^2$, $b=\langle g_A,g_B\rangle$. The first-order change of the resident loss is $\delta(q)=-\eta[(1-q)b+qa]$. (i) $\delta$ is affine in $q$ with slope $-\eta(a-b)$, so reinforcement lowers the first-order resident change whenever $b<a$. (ii) The marginal benefit at $q=0$ is $\eta(a-b)$, larger for antagonistic invaders ($b<0$) and for displaced residents (larger $a$). (iii) For $a>0$, $\delta(q)\le0$ if and only if $q\ge [-b]_+/(a+[-b]_+)$. If $L_A$ is $M_A$-smooth, the exact change is at most $\delta(q)+\tfrac{M_A\eta^2}{2}\|(1-q)g_B+qg_A\|^2$.
\end{proposition}
Part (ii) predicts that replay is most efficient on displacement-prone transitions (E4). The result is local. It does not determine a global replay budget, and we do not claim one.

Table~\ref{tab:theory} maps each result to its observable prediction and test. Proofs are in Appendix~\ref{app:proofs}.

\begin{table}[t]
\caption{From theory to evidence. Each result is stated as an observable prediction and tested by a named experiment.}
\label{tab:theory}
\centering
\scriptsize
\setlength{\tabcolsep}{2.2pt}
\begin{tabular}{p{0.9in}p{1.45in}p{0.52in}p{1.55in}p{0.6in}}
\toprule
Result & Observable prediction & Tests & Evidence & Status \\
\midrule
Thm.~\ref{thm:habitat}, Cor.~\ref{cor:floor} & forgetting rises with $\Omega$; $\Omega$ tracks demand & E1, E2, E8 & $r=-0.789$; $-0.804$/$-0.718$; $\Omega$--probe loss $0.767$ & supported \\
Thm.~\ref{thm:coexist} & $\Omega$ separates coexistence from exclusion & E3 & AUC $1.00$ (natural); $0.93$--$0.97$ (84-suite) & supported \\
Thm.~\ref{thm:ident}, Cor.~\ref{cor:range} & $|r|$ falls with coordinate and outcome spread & scope & CIFAR-100: $0.38/0.18/{-0.05}$ vs.\ predicted $0.32/0.30/0.13$ & supported \\
Prop.~\ref{prop:stabilization} & early $>$ late plasticity and turnover & E7 & $4.78$ $[4.26,5.53]$; $0.37$ $[0.11,0.48]$ & supported \\
Prop.~\ref{prop:reinforce} & replay repairs displacement; efficiency grows with displacement & E4 & 20/20 repaired ($\le325$ ex.); $r(\eta,F_0)=0.70$ & supported \\
Scale claim & transition-level $\gg$ feature-level & E6 & $|r|$: $0.72$ vs.\ $\le0.45$ & supported \\
\bottomrule
\end{tabular}
\end{table}

\section{Experimental design}
\label{sec:exp}
The 8 experiments test the predictions of the theory (Table~\ref{tab:suite}). Full protocols are given in Appendix~\ref{app:protocols}.

\paragraph{Tasks and models.} Natural transitions use the five binary Split-CIFAR-10 tasks $\{0,1\},\dots,\{8,9\}$ \citep{krizhevsky2009cifar} and all 20 ordered pairs, with a multi-head (task-incremental) ResNet-18 \citep{he2016resnet,van2019three,vandeven2022three}. Controlled transitions manipulate task relation directly with pixel permutations of MNIST \citep{lecun1998mnist} and rotations of CIFAR-10. Coexistence is tested on the 20 natural transitions, on 60 pure-domain CIFAR-10 transitions (vehicle and animal binary tasks), and on an 84-transition MNIST suite (4 class pairs $\times$ 7 transforms $\times$ 3 seeds). Scope studies use CIFAR-100, a class-incremental construction, and an ImageNet-pretrained ResNet-50 \citep{deng2009imagenet}.

\paragraph{Protocol.} The default optimizer is Adam \citep{kingma2015adam} with learning rate $10^{-3}$ and batch size 128; E7 and the scope study vary it. $\rho_{A\to B}$ is computed on the frozen resident checkpoint \emph{before} any task-$B$ update, from a class-balanced incoming-task sample and shuffled-label references (256 examples and 20 shuffles in the estimator study). Forgetting is $F_{A\to B}=R_{AA}-R_{BA}$, the drop in task-$A$ test accuracy after learning $B$. A transition coexists when forgetting stays below a fixed threshold ($F<0.20$ on natural transitions, $F<0.10$ on the pure-domain construction) or, on the 84-transition suite, when both tasks keep a fixed fraction (0.80--0.99) of their reference accuracy.

\paragraph{Statistics.} Transitions that share an incoming task are not independent. We therefore report pair means over repeats, incoming-task cluster bootstrap intervals \citep{efron1993bootstrap,cameron2008bootstrap}, exact task-label permutation tests over all $5!$ relabelings \citep{mantel1967detection}, leave-one-pair-out (LOPO) prediction with a no-information baseline, and paired-seed bootstrap intervals for interventions. Scale and mechanism tests used pass criteria fixed before the runs.

\begin{table}[t]
\caption{The eight SLT experiments. Units are directed transitions unless noted.}
\label{tab:suite}
\centering
\scriptsize
\setlength{\tabcolsep}{2.1pt}
\begin{tabular}{lp{1.22in}p{1.28in}p{0.62in}p{1.05in}}
\toprule
ID & Question & Setting & Units & Primary statistic \\
\midrule
E1 & Does $\rho$ order displacement? & Split-CIFAR-10, ResNet-18 & $20\times3$ repeats & $r(\rho,F)$ on pair means \\
E2 & Is the order prospective? & same; $\rho$ from resident only & 20, LOPO & held-out $r$, MAE \\
E3 & Is coexistence a phase? & natural; pure-domain; MNIST suite & 20; 60; 84 & rates; ROC AUC \\
E4 & Does replay reinforce? & buffers $0$--$1000$ & 20 & efficiency, threshold vs.\ $\rho$ \\
E5 & Does $\rho$ help an intervention? & CIFAR-10/100, 6 methods & 3 seeds & ACC, BWT \\
E6 & At which scale? & 5 predictors; probe/LV tests & 20 & $|r|$, pre-registered gates \\
E7 & Are there stages? & per-epoch feature statistics & 15 runs & early$-$late contrasts \\
E8 & Controlled relation? & MNIST perm.; CIFAR-10 rot. & 96; 48 & $r$, group-CV $R^2$ \\
\bottomrule
\end{tabular}
\end{table}

\section{Results}
\label{sec:results}
\subsection{Pre-invasion state orders later displacement (E1, E2, E8)}
\label{sec:e1}
\begin{figure}[t]
\centering
\includegraphics[width=\linewidth]{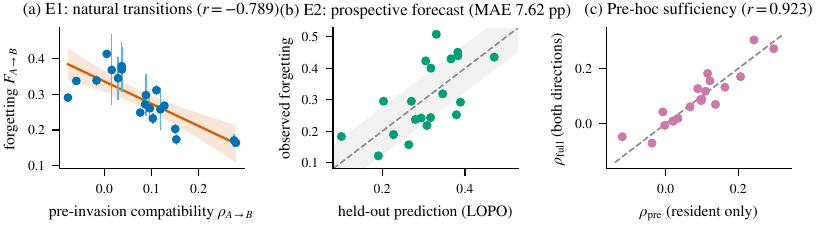}
\caption{\textbf{Compatibility is prospective.} (a) Pair means over three independent repeats (bars: s.d.; band: OLS 95\%). (b) Leave-one-pair-out forecasts (dashed: identity; band: $\pm$MAE). (c) The resident-only score agrees with a score that also uses the reverse direction.}
\label{fig:e1e2}
\end{figure}
Across the three-repeat reproduction, compatibility strongly anticorrelates with later forgetting (Figure~\ref{fig:e1e2}a): $r=-0.789$, Spearman $-0.789$, $R^2=0.622$. Resampling incoming tasks gives a 95\% interval of $[-0.90,-0.72]$, and the association is negative within four of five incoming tasks. It holds in every repeat on its own ($r=-0.706$, $-0.720$, $-0.674$), across all 60 individual runs ($r=-0.695$), and in an independent single-run reproduction ($r=-0.710$, $p=4.6\times10^{-4}$). The coordinate is highly repeatable (mean between-repeat correlation $0.96$ for $\rho$ and $0.77$ for forgetting), and a separate revalidation with a task-label permutation test gives $r=-0.723$, $p=0.033$. Forgetting spans $0.16$--$0.41$ (mean $0.285$), so the ordering is not produced by a few extreme pairs.

The same ordering is available before adaptation. Leave-one-pair-out forecasts reach $r=0.648$ with MAE $7.62$ percentage points, against $10.03$ pp for a no-information predictor that uses the mean of the other pairs, a 24\% error reduction (Figure~\ref{fig:e1e2}b). A stricter split that holds out every transition into the same incoming task gives $r=0.669$ and MAE $7.44$ pp. A resident-only score agrees with a score that also uses the reverse direction ($r=0.923$; Figure~\ref{fig:e1e2}c) and loses little predictive power ($r=-0.753$ vs.\ $-0.771$ with forgetting).

\begin{figure}[t]
\centering
\begin{minipage}[t]{0.40\linewidth}
\centering
\includegraphics[width=\linewidth]{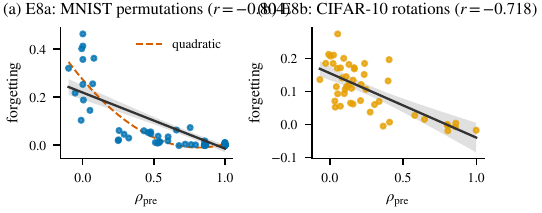}
\end{minipage}\hfill
\begin{minipage}[t]{0.58\linewidth}
\centering
\includegraphics[width=\linewidth]{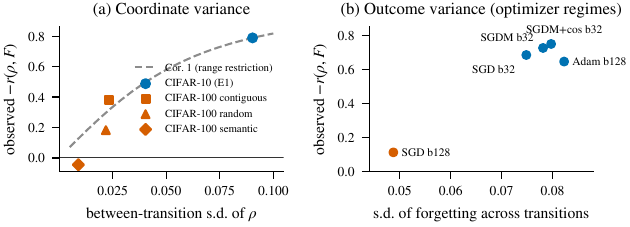}
\end{minipage}
\caption{\textbf{Left:} controlled manipulations of task relation (E8); black line and band: linear fit with 95\% interval; red dashed: quadratic fit on MNIST. \textbf{Right:} identifiability (Theorem~\ref{thm:ident}). (a) Predictive strength falls with the between-transition spread of $\rho$ along the range-restriction curve of Corollary~\ref{cor:range}, anchored at E1. (b) Across optimizer regimes, the one regime with collapsed outcome variance is the one where ordering vanishes.}
\label{fig:controlled}
\end{figure}

Controlled manipulations recover the same direction (Figure~\ref{fig:controlled}, left). MNIST permutations give $r=-0.804$ over 96 transitions (group cross-validated $R^2=0.626$) and CIFAR-10 rotations give $r=-0.718$ over 48 (group-CV $R^2=0.484$). On MNIST, a quadratic term raises group-CV $R^2$ to $0.764$, with a minimum near $\rho\approx0.81$. Forgetting reaches a floor once the invader is almost fully supported. On CIFAR-10 the quadratic term adds nothing ($0.465$).

A bridge analysis relates the empirical coordinate to the demand in Theorem~\ref{thm:habitat} without identifying the two. Within incoming task, $\Omega$ correlates $0.767$ with the unclipped incoming probe loss (cluster 95\% CI $[0.28,0.95]$) and $0.692$ with the clipped deficit $H$ (CI $[-0.14,0.94]$ with five clusters; Appendix~\ref{app:bridge}). The unclipped link is the robust one.

\subsection{Coexistence is a phase of the same transition (E3)}
\label{sec:e3}
\begin{figure}[t]
\centering
\includegraphics[width=\linewidth]{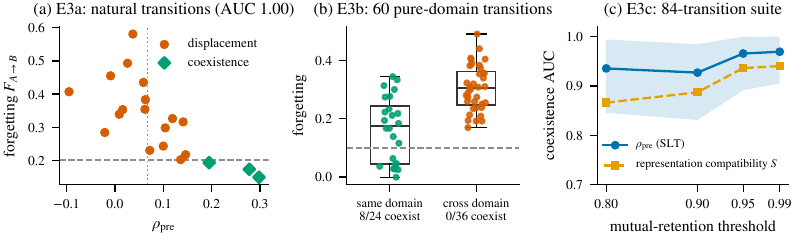}
\caption{\textbf{Coexistence is a phase of the transition.} (a) Natural transitions: the three that coexist (green, $F<0.20$) are the three most compatible; dotted line: compatibility split. (b) Pure-domain construction: only same-domain transitions reach near-zero forgetting. (c) 84-transition suite: coexistence AUC of $\rho$ and of a representation-compatibility score $S$ across mutual-retention thresholds (band: 95\% cluster bootstrap over 28 class-pair$\times$transform clusters).}
\label{fig:coexist}
\end{figure}
Corollary~\ref{cor:floor} and Theorem~\ref{thm:coexist} predict a phase boundary. Compatible invaders can establish without displacing the resident, while incompatible ones cannot. The reproduced natural transitions follow this pattern (Figure~\ref{fig:coexist}a). Three of the 20 transitions coexist ($F<0.20$), and they are the three with the highest compatibility (AUC $1.00$). Splitting at $\rho=0.067$, compatible transitions forget $0.235$ on average and coexist in 30\% of cases, whereas incompatible ones forget $0.408$ and never coexist (Mann--Whitney $p=2.2\times10^{-4}$). A stricter construction within semantic domains gives the same pattern. 8 of 24 same-domain transitions reach near-zero forgetting against 0 of 36 cross-domain transitions, with mean forgetting $0.166$ versus $0.306$ ($p=1.25\times10^{-5}$; Figure~\ref{fig:coexist}b). On the 84-transition suite, where compatibility spans a wide range, $\rho$ separates mutual coexistence from exclusion at every retention threshold, with AUC $0.936$, $0.927$, $0.966$, and $0.969$ at 0.80, 0.90, 0.95, and 0.99 (Figure~\ref{fig:coexist}c; at 0.90, 95\% CI $[0.83,0.98]$). It exceeds the representation-compatibility score $S$ at all four thresholds, most clearly at 0.80 ($+0.070$, CI $[0.001,0.169]$), keeps AUC $0.903$ after removing the 12 identity transitions, and correlates $-0.82$ with forgetting in this suite (Appendix~\ref{app:coexist}).

\subsection{The useful ecological scale is mesoscopic (E6)}
\label{sec:e6}
\begin{figure}[t]
\centering
\includegraphics[width=0.80\linewidth]{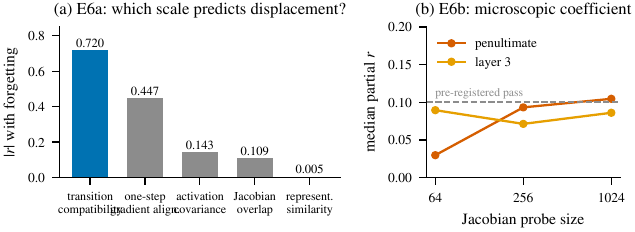}
\caption{\textbf{Scale selection.} (a) On the same 20 transitions, the transition-level coordinate predicts displacement far better than feature- or step-level quantities. (b) Increasing Jacobian probes 16-fold at two layers does not produce a predictive microscopic coefficient.}
\label{fig:scale}
\end{figure}
We next test whether a literal microscopic ecological model is needed. On the same 20 natural transitions, compatibility has $|r|=0.720$ with forgetting, compared with $0.447$ for one-step gradient alignment, $0.143$ for activation covariance, $0.109$ for raw Jacobian overlap, and $0.005$ for representation similarity (Figure~\ref{fig:scale}a). Feature-level competition coefficients fail pre-registered tests in a regime with substantial forgetting (s.d.\ $0.064$): task-level $r=0.177$ (exact permutation $p=0.54$); increasing Jacobian probes from 64 to 1024 leaves the median partial correlation at or below $0.105$ (Figure~\ref{fig:scale}b); fixed-coefficient Lotka--Volterra fits beat simpler baselines in 0 of 60 transitions (median held-out $R^2=-29.2$); none of 330 local windows has stable coefficients; and neural $R^\ast$ resources are identifiable for only 1.9\% of candidate directions (Appendix~\ref{app:scale}). These results support \emph{scale selection}. The ecological structure is predictive after coarse-graining to a resident--invader transition, while the tested feature-level coefficients do not behave as stationary population parameters.

\subsection{Learning exhibits successional stages (E7)}
\label{sec:e7}
\begin{figure}[t]
\centering
\includegraphics[width=\linewidth]{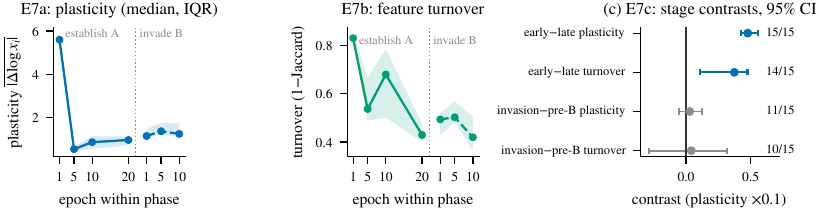}
\caption{\textbf{Establishment precedes stabilization.} (a, b) Median and interquartile range over 15 runs of per-epoch plasticity and turnover of the top-quartile resident features, while $A$ is established and then while $B$ invades. (c) Pre-specified contrasts with 95\% intervals; grey: interval includes zero.}
\label{fig:stages}
\end{figure}
The succession account also makes a temporal prediction. Across 15 runs, early-minus-late plasticity is $4.776$ (95\% CI $[4.262,5.532]$, 15/15 positive) and early-minus-late feature turnover is $0.374$ (CI $[0.107,0.475]$, 14/15 positive; Figure~\ref{fig:stages}). Large early reorganization gives way to a more stable regime, consistent with Proposition~\ref{prop:stabilization}. The invasion does not reliably reopen a pioneer phase. invasion-minus-pre-$B$ contrasts are positive in 11/15 and 10/15 runs but their intervals include zero, so we do not claim renewed establishment after invasion.

\subsection{Reinforcement and interventions (E4, E5)}
\label{sec:e4}
Replay reinforces the resident in every transition. Without replay, mean forgetting is $0.289$. All 20 transitions are brought within 10\% of their resident accuracy using at most 325 stored examples (median 100). Proposition~\ref{prop:reinforce}(ii) predicts that the per-example benefit grows with the displacement being repaired: initial replay efficiency correlates $0.70$ with forgetting at zero replay and $-0.42$ with compatibility (Spearman $-0.46$), so the most incompatible transitions gain most from their first replayed examples (Appendix~\ref{app:replay}). The exact buffer at which a transition crosses the 10\% line is not ordered by $\rho$ ($r=0.01$), consistent with the local scope of Proposition~\ref{prop:reinforce}. The coordinate also carries over to interventions (E5; Appendix~\ref{app:intervention}). Combined with replay, compatibility-modulated regularization outperforms fine-tuning, EWC, and A-GEM on both Split-CIFAR-10 and Split-CIFAR-100 (ACC $0.815$ and $0.531$), and gating EWC-DR by $\Omega$ raises mean ACC from $0.800$ to $0.824$ while cutting forgetting by 40\% (three seeds). DER++ remains the strongest single method, so SLT is best used as a diagnostic layer that tells such methods where protection is needed.

\subsection{Robustness and scope}
\label{sec:scope}
\begin{table}[t]
\caption{Main results with uncertainty. CI: 95\% interval (cluster bootstrap unless stated).}
\label{tab:main}
\centering
\scriptsize
\setlength{\tabcolsep}{2.3pt}
\begin{tabular}{p{1.62in}p{1.5in}p{2.05in}}
\toprule
Claim & Result & Uncertainty / comparison \\
\midrule
E1 natural ordering & $r=-0.789$, $R^2=0.622$ & CI $[-0.90,-0.72]$; each repeat $\le-0.67$; single run $-0.710$ \\
E2 held-out forecast & MAE $7.62$ pp, $r=0.648$ & no-information MAE $10.03$ pp; LOTO $r=0.669$ \\
E8 controlled relation & $r=-0.804$ / $-0.718$ & group-CV $R^2$ $0.626$ / $0.484$ \\
E3 coexistence phase & AUC $1.00$ / $0.927$ & natural / 84-suite (CI $[0.83,0.98]$); 8/24 vs.\ 0/36 \\
E4 reinforcement & 20/20 repaired & $\le325$ examples (median 100); $r(\eta,F_0)=0.70$ \\
E5 intervention & ACC $0.815$ / $0.531$ & above fine-tune, EWC, A-GEM on both benchmarks \\
E6 transition vs.\ micro scale & $|r|=0.720$ & gradient $0.447$; others $\le0.143$; LV 0/60 \\
E7 successional stages & $+4.78$ plasticity, $+0.37$ turnover & CI $[4.26,5.53]$, $[0.11,0.48]$ \\
\bottomrule
\end{tabular}
\end{table}
The ordering persists across estimator settings. all 24 probe-size and shuffle combinations give $r$ between $-0.833$ and $-0.662$, whereas CKA \citep{kornblith2019cka} gives $+0.113$ and a plain linear-probe accuracy $-0.464$ (Table~\ref{tab:main}). The score needs few labels, but the tested label-free variants do not recover the signal. 10 labels per class correlate $0.816$ with the full estimate, while two label-free variants carry no signal. On an ImageNet-pretrained ResNet-50 the direction holds ($r=-0.426$, 10/10 negative leave-one-task-out slopes) with a wide cluster interval. Details are in Appendix~\ref{app:robust}.

Theorem~\ref{thm:ident} predicts where a prospective coordinate should lose leverage. The observed pattern follows this prediction (Figure~\ref{fig:controlled}, right). On three CIFAR-100 partitions of ten 10-class tasks, the spread of $\rho$ falls from $0.090$ to $0.023$, $0.022$, and $0.009$, and $-r$ falls from $0.789$ to $0.381$, $0.182$, and $-0.045$. Anchored at E1, Corollary~\ref{cor:range} predicts $0.32$, $0.30$, and $0.13$, matching both the ordering and the size of the contraction. Across five optimizer regimes, four give $r$ between $-0.646$ and $-0.749$; the fifth (plain SGD, batch 128) is the one whose forgetting spread collapses ($0.049$ vs.\ $\ge0.075$), and there the ordering vanishes ($r=-0.111$), again as the theorem requires.

\section{Discussion}
The evidence supports a task-scale claim rather than a literal ecological simulator of neural features. Continual-learning transitions show reproducible \emph{successional structure} at this scale. Before adaptation, directional compatibility orders how much the resident will be displaced and resolves a coexistence/exclusion phase; replay repairs displacement where it is largest; during adaptation, plasticity and turnover fall from establishment toward stabilization. The scale result distinguishes SLT from an ecological relabeling of task similarity. The predictive abstraction is a transition state, not a static pairwise coefficient or a neuron-level resource law. The identifiability law specifies when that state should be measurable. Accuracy-oriented methods determine how to preserve knowledge. SLT instead diagnoses the upcoming transition from a frozen model and a small labelled sample, so it can be used alongside replay, regularization, projection, and isolation methods.

\paragraph{Limitations.} The evidence comes from vision task-incremental settings with explicit task boundaries and a small labelled sample of the incoming task; wide tasks (CIFAR-100) and single-head class-incremental learning compress the between-transition signal, as Theorem~\ref{thm:ident} anticipates. $\sigma_{A\to B}$ and $s^\star_{A\to B}$ are latent, so E3 validates the phase rather than the quantities, and within one semantic construction domain membership carries information beyond $\rho$ (Appendix~\ref{app:coexist}). The coordinate characterizes transitions that differ in task relation. Replay is described by a local principle rather than a global budget, and Proposition~\ref{prop:stabilization} is a sufficient idealized result. SLT is a diagnostic framework and is not proposed as a new accuracy-leading continual-learning algorithm.

\section{Conclusion}
SLT treats continual learning as a sequence of invasions into already-occupied representations. The transition can be diagnosed before adaptation, ends in coexistence or displacement, responds to resident reinforcement, and follows an establishment-to-stabilization path; the identifiability law predicts where this mesoscopic description stops working.

\section*{AI Use Statement}
Generative AI tools were used for literature exploration, code assistance, statistical cross-checking, figure preparation, and language editing. The authors verified mathematical statements, experimental procedures, numerical claims, citations, code, and final prose against executed analyses and primary sources. The authors are responsible for the submission.

\section*{Ethics Statement}
This work uses standard public image-classification benchmarks and does not involve human subjects, private data, or deployment decisions.

\section*{Reproducibility Statement}
The code repository at \url{https://github.com/shoaibdipu/Neural_Succession/} contains the training code for all experiments, the raw result records and reproduction logs, and one analysis script, \texttt{scripts/make\_figures\_and\_stats.py}, which regenerates every figure and every derived statistic (bootstrap intervals, permutation tests, held-out baselines, range-restriction predictions) into \texttt{data/derived\_revision\_stats.json}. Appendix~\ref{app:protocols} gives per-experiment protocols.

\bibliography{refs}
\bibliographystyle{paperstyle}

\newpage
\appendix
\section{Proofs}
\label{app:proofs}
Throughout, $\|\cdot\|$ is the Euclidean norm on features and logits and the operator norm on linear read-outs; $\mathbb E_B$ and $\mathbb E_A$ denote expectations over the incoming and resident task distributions.

\subsection{Proof of Theorem~\ref{thm:habitat}}
\textbf{Setting.} A read-out $w$ maps features to logits, $z=w\,\phi(x)$, and the loss satisfies $|\ell(z,y)-\ell(z',y)|\le\lambda_B\|z-z'\|$ for all $z,z',y$. For a representation $\phi$ write $\mathcal R(\phi)=\mathcal R_B^{(R_B)}(\phi)=\inf_{\|w\|\le R_B}\mathbb E_B\,\ell(w\phi(x),y)$.

\textbf{Step 1 (near-optimal probe for the target).} Fix $\phi\in\mathcal E_B$, so $\mathcal R(\phi)\le\ell_B^\star$. For any $\epsilon>0$ choose $w_\epsilon$ with $\|w_\epsilon\|\le R_B$ and $\mathbb E_B\ell(w_\epsilon\phi(x),y)\le\mathcal R(\phi)+\epsilon\le\ell_B^\star+\epsilon$.

\textbf{Step 2 (transport the probe to the resident).} Because $w_\epsilon$ is feasible for the resident representation,
\[
\mathcal R(\phi_A)\le\mathbb E_B\ell(w_\epsilon\phi_A(x),y)
\le \mathbb E_B\ell(w_\epsilon\phi(x),y)+\lambda_B\,\mathbb E_B\|w_\epsilon(\phi_A(x)-\phi(x))\|
\le \ell_B^\star+\epsilon+\lambda_BR_B\,\mathbb E_B\|\phi_A(x)-\phi(x)\|,
\]
using Lipschitzness in the logits and $\|w_\epsilon v\|\le\|w_\epsilon\|\|v\|$.

\textbf{Step 3 (conclude).} Jensen's inequality gives $\mathbb E_B\|\phi_A-\phi\|\le(\mathbb E_B\|\phi_A-\phi\|^2)^{1/2}=D_B(\phi)$. Hence $\mathcal R(\phi_A)-\ell_B^\star\le\epsilon+\lambda_BR_BD_B(\phi)$ for every $\epsilon>0$, so $\mathcal R(\phi_A)-\ell_B^\star\le\lambda_BR_BD_B(\phi)$. Since $D_B(\phi)\ge0$, the same bound holds for the positive part, $H_{A\to B}\le\lambda_BR_BD_B(\phi)$, which is the claim. If $H_{A\to B}=0$ the statement is trivial. $\square$

\subsection{Proof of Corollary~\ref{cor:floor}}
If $H_{A\to B}=0$ then $\Pi_{A\to B}=0$ and both claims are immediate. Otherwise Theorem~\ref{thm:habitat} gives $D_B(\phi)\ge H_{A\to B}/(\lambda_BR_B)>0$ for every $\phi\in\mathcal E_B$, so every establishing representation lies in the set over which $\sigma_{A\to B}$ is defined and satisfies $S_A(\phi)\ge\sigma_{A\to B}D_B(\phi)$. Combining the two inequalities, $S_A(\phi)\ge\sigma_{A\to B}H_{A\to B}/(\lambda_BR_B)=\Pi_{A\to B}$. The lower side of Assumption~\ref{ass:sens} then gives $[L_A(\phi)-L_A(\phi_A)]_+\ge\frac{m_A}{2}S_A(\phi)^2\ge\frac{m_A}{2}\Pi_{A\to B}^2$. Finally, if $\Pi_{A\to B}>\sqrt{2\delta_A/m_A}$ then $\frac{m_A}{2}\Pi_{A\to B}^2>\delta_A$, so every $B$-establishing representation raises the resident loss by more than $\delta_A$. $\square$

\subsection{Proof of Theorem~\ref{thm:coexist}}
Let $g(t)=\frac{M_A}{2}(s^\star_{A\to B}+t)^2$ for $t\ge0$. By hypothesis $g(0)\le\delta_A$, and $g$ is continuous and increasing, so for any $\epsilon>0$ there is $t_\epsilon>0$ with $g(t_\epsilon)\le\delta_A+\epsilon$. By definition of the infimum there is $\phi_\epsilon\in\mathcal E_B$ with $S_A(\phi_\epsilon)\le s^\star_{A\to B}+t_\epsilon$. This $\phi_\epsilon$ establishes $B$ (it lies in $\mathcal E_B$), and the upper side of Assumption~\ref{ass:sens} gives $[L_A(\phi_\epsilon)-L_A(\phi_A)]_+\le\frac{M_A}{2}S_A(\phi_\epsilon)^2\le g(t_\epsilon)\le\delta_A+\epsilon$. If the infimum is attained by some $\phi^\star\in\mathcal E_B$, the same argument with $t=0$ gives an increase of at most $g(0)\le\delta_A$. $\square$

\subsection{Proof of Theorem~\ref{thm:ident}}
Write $\sigma_\Omega^2=\operatorname{Var}(\Omega)$. Because $\mathbb E[\varepsilon\mid\Omega]=0$, $\operatorname{Cov}(\Omega,\varepsilon)=\mathbb E[(\Omega-\mathbb E\Omega)\,\mathbb E[\varepsilon\mid\Omega]]=0$. Hence
\[
\operatorname{Cov}(F,\Omega)=\beta\sigma_\Omega^2,\qquad \operatorname{Var}(F)=\beta^2\sigma_\Omega^2+\sigma_\varepsilon^2,\qquad
R^2(F\mid\Omega)=\frac{\operatorname{Cov}(F,\Omega)^2}{\operatorname{Var}(F)\operatorname{Var}(\Omega)}=\frac{\beta^2\sigma_\Omega^2}{\beta^2\sigma_\Omega^2+\sigma_\varepsilon^2}.
\]
For the measured coordinate $\widetilde\Omega=\Omega+\eta$ with $\eta$ independent of $(\Omega,\varepsilon)$ and variance $\sigma_\eta^2$, $\operatorname{Cov}(F,\widetilde\Omega)=\operatorname{Cov}(F,\Omega)=\beta\sigma_\Omega^2$ and $\operatorname{Var}(\widetilde\Omega)=\sigma_\Omega^2+\sigma_\eta^2$. Therefore
\[
\operatorname{Corr}(F,\widetilde\Omega)^2=\frac{\beta^2\sigma_\Omega^4}{(\beta^2\sigma_\Omega^2+\sigma_\varepsilon^2)(\sigma_\Omega^2+\sigma_\eta^2)}=R^2(F\mid\Omega)\cdot\frac{\sigma_\Omega^2}{\sigma_\Omega^2+\sigma_\eta^2},
\]
which is the stated attenuation. Both factors lie in $[0,1]$, so predictive power is limited by the between-transition spread of the coordinate relative to outcome noise and to measurement noise. $\square$

\subsection{Proof of Corollary~\ref{cor:range}}
In the first ecology, Theorem~\ref{thm:ident} gives $r^2=\beta^2\sigma_\Omega^2/(\beta^2\sigma_\Omega^2+\sigma_\varepsilon^2)$; if $r\ne0$ this rearranges to $\sigma_\varepsilon^2=\beta^2\sigma_\Omega^2(1-r^2)/r^2$. In the second ecology the slope and noise are unchanged and the spread is $s\sigma_\Omega$, so
\[
r'^2=\frac{\beta^2s^2\sigma_\Omega^2}{\beta^2s^2\sigma_\Omega^2+\beta^2\sigma_\Omega^2(1-r^2)/r^2}=\frac{s^2r^2}{s^2r^2+1-r^2}.
\]
Taking square roots, $|r'|=|r|s/\sqrt{1-r^2+r^2s^2}$, and $r'$ has the sign of $\beta$. The map $s\mapsto|r'|$ is increasing, equals $|r|$ at $s=1$, and tends to $0$ as $s\to0$. $\square$

\subsection{Proof of Proposition~\ref{prop:stabilization}}
Let $\Delta(t)=L(\theta_t)-L^\ast\ge0$. Along gradient flow, $\dot\theta_t=-\nabla L(\theta_t)$ and $\dot\Delta=-\|\nabla L(\theta_t)\|^2$. The PL inequality $\frac12\|\nabla L(\theta)\|^2\ge\mu(L(\theta)-L^\ast)$ gives $\dot\Delta\le-2\mu\Delta$, hence $\Delta(t)\le e^{-2\mu t}\Delta(0)$ by Gr\"onwall's inequality.

\textbf{Finite path length.} Whenever $\Delta(t)>0$,
\[
\frac{d}{dt}\sqrt{\Delta(t)}=-\frac{\|\nabla L(\theta_t)\|^2}{2\sqrt{\Delta(t)}}\le-\sqrt{\frac{\mu}{2}}\,\|\nabla L(\theta_t)\|,
\]
using $\|\nabla L\|\ge\sqrt{2\mu\Delta}$. Integrating from $t$ to $T$ gives $\int_t^T\|\dot\theta_s\|\,ds\le\sqrt{2/\mu}\,\big(\sqrt{\Delta(t)}-\sqrt{\Delta(T)}\big)\le\sqrt{2\Delta(t)/\mu}$. If $\Delta$ reaches $0$ at a finite time, the gradient vanishes there and the trajectory stops, so the bound still holds. Letting $T\to\infty$, $\int_t^\infty\|\dot\theta_s\|\,ds\le\sqrt{2\Delta(0)/\mu}\,e^{-\mu t}<\infty$.

\textbf{Feature variation.} By the chain rule and the gradient bound on active features, $|\tfrac{d}{dt}\log x_i(\theta_t)|=|\langle\nabla_\theta\log x_i(\theta_t),\dot\theta_t\rangle|\le C_i\|\dot\theta_t\|$. Therefore
\[
\int_t^\infty\Big|\frac{d}{dt}\log x_i(\theta_s)\Big|\,ds\le C_i\sqrt{\frac{2\Delta(0)}{\mu}}\,e^{-\mu t},
\]
which is finite for $t=0$ and decays exponentially in $t$: all but an exponentially small part of each feature's total change occurs early, which is the establishment-then-stabilization profile tested in E7. Smoothness is not needed for this argument; it only guarantees that gradient flow is well defined. $\square$

\subsection{Proof of Proposition~\ref{prop:reinforce}}
The step on $L_q=(1-q)L_B+qL_A$ is $\Delta\theta=-\eta[(1-q)g_B+qg_A]$. The first-order change of the resident loss is $\langle g_A,\Delta\theta\rangle=-\eta[(1-q)\langle g_A,g_B\rangle+q\|g_A\|^2]=-\eta[(1-q)b+qa]=\delta(q)$.
(i) $\delta$ is affine in $q$ with $\delta'(q)=-\eta(a-b)$, which is negative exactly when $b<a$; increasing $q$ then lowers the first-order change.
(ii) The marginal benefit of reinforcement is $-\delta'(q)=\eta(a-b)$. It increases with $a=\|g_A\|^2$, which is large when the resident has been displaced from its optimum, and decreases with $b$, so it is largest for antagonistic invaders with $b<0$.
(iii) $\delta(q)\le0$ iff $q(a-b)\ge-b$. If $b\ge0$ this holds for every $q\in[0,1]$. If $b<0$, then $a-b>0$ and the condition is $q\ge-b/(a-b)=[-b]_+/(a+[-b]_+)$, which lies in $(0,1]$.
For the exact change, if $\nabla L_A$ is $M_A$-Lipschitz then $L_A(\theta+\Delta\theta)=L_A(\theta)+\int_0^1\langle\nabla L_A(\theta+t\Delta\theta),\Delta\theta\rangle dt\le L_A(\theta)+\langle g_A,\Delta\theta\rangle+\frac{M_A}{2}\|\Delta\theta\|^2$ (the descent lemma), which gives the stated bound with $\|\Delta\theta\|^2=\eta^2\|(1-q)g_B+qg_A\|^2$. $\square$

\section{Experimental protocols}
\label{app:protocols}
\paragraph{E1 natural transitions.} Five binary Split-CIFAR-10 tasks; all 20 ordered pairs; task-incremental ResNet-18 with one head per task. For each pair, task $A$ is learned, $\rho_{A\to B}$ is computed on the frozen checkpoint, task $B$ is learned, and $F_{A\to B}$ is recorded. The whole pipeline is repeated three times and analyzed on pair means. The task-label permutation test permutes the identities of the five tasks in the $\rho$ matrix, keeping the forgetting matrix fixed, and recomputes the correlation over the 20 off-diagonal entries for all 120 relabelings. On the three-repeat pair means this test gives one-sided $p=0.067$ (8 of 120 relabelings at least as extreme; with five tasks the smallest attainable value is $1/120$). The separate revalidation used the permutation-invariant $\rho$ with 256 probe examples and 20 shuffles. The single-run reproduction repeats the whole pipeline once with a new seed.
\paragraph{E2 prospective forecast.} For each of the 20 pairs, a linear model of forgetting on $\rho$ is fit to the other 19 and used to predict the held-out pair. The no-information baseline predicts the mean forgetting of the other 19. E1.5 compares the resident-only score with a score that also uses the reverse direction.
\paragraph{E3 coexistence.} (a) The 20 natural E1 transitions; coexistence when $F<0.20$; transitions are split at $\rho=0.067$ into 10 compatible and 10 incompatible. (b) Binary tasks built within a single semantic domain of CIFAR-10 (vehicles: airplane, automobile, ship, truck; animals: the remaining six classes); 24 same-domain and 36 cross-domain directed transitions; coexistence when $F<0.10$. (c) MNIST suite: class pairs $\{0/1, 2/7, 3/5, 4/9\}$ under identity, rotations of $15^\circ,45^\circ,90^\circ$, and pixel permutations of 30\%, 70\%, 100\%, three seeds, 84 transitions; mutual coexistence at threshold $\tau$ when the resident keeps at least a fraction $\tau$ of its accuracy and the invader reaches the same fraction of its reference accuracy, $\tau\in\{0.80,0.90,0.95,0.99\}$. Intervals resample the 28 class-pair$\times$transform clusters.
\paragraph{E4 replay.} Same 20 transitions; replay buffers from 0 to 1000 resident examples in steps of 25, mixed into $B$ training. $M^\ast$ is the smallest buffer reducing forgetting below $0.1\,R_{AA}$; efficiency $\eta$ is the forgetting reduction per stored example between buffers 0 and 25.
\paragraph{E5 interventions.} Six methods on Split-CIFAR-10 and Split-CIFAR-100 (task-incremental, three seeds): fine-tuning, EWC, A-GEM, DER++, compatibility-weighted EWC, and compatibility-weighted EWC with replay. The gated test compares an EWC regularization baseline (EWC-DR) against the same method with its regularization strength multiplied by the mean of $\Omega$ from the resident tasks to the incoming task, and a gradient-gated replay against uniform replay at the matched mean weight. The pre-specified pass rule was $+1$ pp ACC with BWT not worse.
\paragraph{E6 scale.} On the 20 E1 transitions: activation covariance, centered representation similarity, raw Jacobian overlap, and one-step gradient alignment, each measured before $B$ training. Mechanism tests (identifiable regime: plain SGD, lr 0.03, batch 32) evaluate Jacobian-derived competition coefficients at the feature and task levels, fixed-coefficient and locally stationary Lotka--Volterra fits against constant-rate and shuffled-coefficient baselines, a 64/256/1024-probe by two-layer sweep, and $R^\ast$ resource identification.
\paragraph{E7 stages.} Fifteen runs (five transitions, three seeds). Feature statistics are recorded at epochs $\{0,1,5,10,20\}$ of task $A$ and $\{0,1,5,10\}$ of task $B$. Active features are the top quartile by resident strength; plasticity is the mean absolute change of log-strength between checkpoints; turnover is one minus the Jaccard overlap of active sets. Early/late contrasts compare the first and last checkpoint intervals of task-$A$ training; invasion contrasts compare the first interval of $B$ with the last interval of $A$.
\paragraph{E8 controlled.} MNIST: fixed pixel permutations at graded fractions (96 transitions). CIFAR-10: rotations at graded angles (48 transitions). Group cross-validation holds out whole manipulation groups.
\paragraph{Robustness and scope.} Estimator grid: $\{32,64,128,256,512,1000\}$ probe examples $\times$ $\{1,5,20,100\}$ shuffles. Label budgets: 2--100 labels per class, plus two label-free variants. Optimizers: Adam (lr $10^{-3}$, batch 128), SGD (lr 0.03, batch 32 or 128), SGD with momentum (lr 0.03, batch 32), and SGD with momentum and cosine schedule (lr 0.1, batch 32). Pretrained: ImageNet ResNet-50. CIFAR-100: ten 10-class tasks under contiguous, random, and superclass-based partitions (90 directed transitions each).

\section{Additional results}
\subsection{Coexistence details}
\label{app:coexist}
\begin{table}[H]
\caption{E3c: coexistence AUC on the 84-transition suite (95\% cluster-bootstrap intervals).}
\centering
\scriptsize
\begin{tabular}{ccccc}
\toprule
Threshold $\tau$ & Coexistence rate & $\rho_{\mathrm{pre}}$ (SLT) & Representation compatibility $S$ & Difference \\
\midrule
0.80 & 0.667 & 0.936 $[0.85,0.99]$ & 0.866 $[0.74,0.96]$ & $+0.070$ $[0.001,0.169]$ \\
0.90 & 0.571 & 0.927 $[0.83,0.98]$ & 0.887 $[0.76,0.97]$ & $+0.040$ $[-0.014,0.113]$ \\
0.95 & 0.476 & 0.966 $[0.89,1.00]$ & 0.936 $[0.81,1.00]$ & $+0.030$ $[-0.005,0.106]$ \\
0.99 & 0.310 & 0.969 $[0.91,1.00]$ & 0.940 $[0.85,0.99]$ & $+0.029$ $[-0.026,0.112]$ \\
\bottomrule
\end{tabular}
\end{table}
Excluding the 12 identity transitions, AUC at $\tau=0.90$ is $0.903$ for $\rho$ and $0.850$ for $S$. In the pure-domain construction (E3b) the phase separation follows domain membership; within that construction $\rho$ is only weakly related to forgetting ($r=-0.19$), so task relation beyond $\rho$ contributes to coexistence there.

\subsection{Establishment bridge}
\label{app:bridge}
Theorem~\ref{thm:habitat} is stated for the establishment deficit $H_{A\to B}$, while the experiments measure $\Omega$. The bridge analysis tests whether the two move together. For each Split-CIFAR-10 transition we take $\Omega$ from the estimator study (256 probe examples, 20 shuffles) and the incoming probe loss and clipped deficit from probes fitted on the frozen resident; both are averaged over three seeds. Because incoming tasks differ in difficulty, we remove each incoming task's mean before correlating, so the comparison is between residents facing the same invader. Within incoming task, $\Omega$ tracks the probe loss closely and the clipped deficit more loosely, because clipping at zero removes variation among well-supported transitions. On semantic CIFAR-100 the coordinate has almost no spread and the link disappears, which is the same contraction that Theorem~\ref{thm:ident} predicts for the behavioral ordering.
\begin{figure}[H]
\centering
\includegraphics[width=0.62\linewidth]{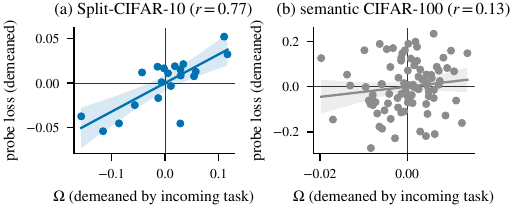}
\caption{Establishment bridge. Within incoming task, $\Omega$ tracks the incoming probe loss on Split-CIFAR-10 ($r=0.767$, cluster 95\% CI $[0.28,0.95]$; clipped deficit $H$: $0.692$, $[-0.14,0.94]$ with five clusters) but not on semantic CIFAR-100 ($r=0.131$), where the coordinate has little spread.}
\end{figure}

\subsection{Replay}
\label{app:replay}
Each of the 20 transitions is retrained with resident buffers from 0 to 1000 examples in steps of 25. Forgetting falls with buffer size in every transition, and all 20 reach the 10\% criterion within 325 examples (median 100). The initial efficiency $\eta$, the drop in forgetting per stored example over the first 25 examples, is largest where displacement without replay is largest ($r=0.70$) and on incompatible transitions ($r=-0.42$, Spearman $-0.46$). Proposition~\ref{prop:reinforce}(ii) predicts both relations, since the marginal value of reinforcement grows with the resident gradient that displacement creates. The buffer at which the criterion is first met is not ordered by $\rho$, which fits the local character of the proposition: it describes where each additional replayed example helps most, not a universal budget.
\begin{figure}[H]
\centering
\includegraphics[width=\linewidth]{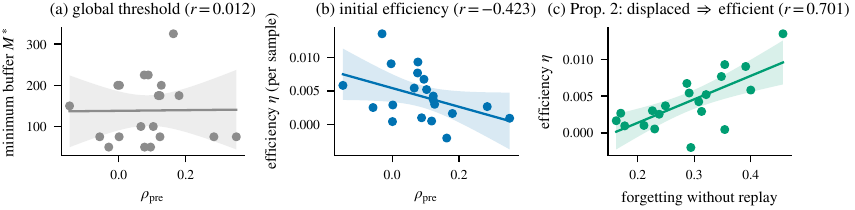}
\caption{Replay (E4). (a) The global minimum buffer is not ordered by $\rho$. (b) Initial per-sample efficiency decreases with $\rho$. (c) Efficiency rises with forgetting at zero replay, as Proposition~\ref{prop:reinforce}(ii) predicts; part of this relation is expected from headroom.}
\end{figure}

\subsection{Interventions}
\label{app:intervention}
The benchmark places compatibility-aware variants among standard task-incremental methods trained with the same backbone and schedule. Adding replay to compatibility-modulated regularization lifts it above fine-tuning, EWC, and A-GEM on both datasets; modulation alone helps on CIFAR-10 (ACC $0.569$ vs.\ $0.526$ for EWC) but not on CIFAR-100, and DER++ remains the strongest single method. In a separate protection study on the 20 Split-CIFAR-10 transitions (three seeds), every transition gains from protection, and the gain rises with incompatibility: across pair means the reduction in forgetting correlates $0.69$ with $\Omega$ for DER++ (mean gain $0.162$) and $0.55$ for EWC-DR (mean gain $0.216$). This is the pattern Proposition~\ref{prop:reinforce} predicts, since incompatible transitions displace the resident most and so leave the most to recover. The gated sign tests below compare each gate with its ungated counterpart at matched budget.
\begin{figure}[H]
\centering
\includegraphics[width=\linewidth]{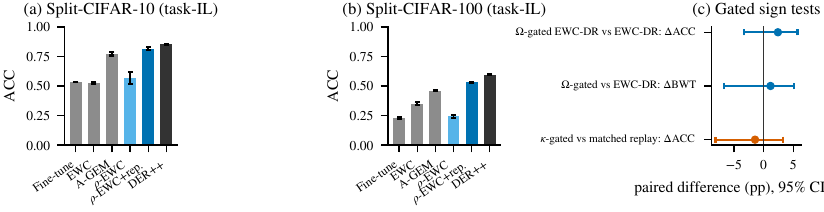}
\caption{Interventions (E5). (a, b) Task-incremental accuracy (mean $\pm$ s.d., three seeds). (c) Paired differences for the gated sign tests.}
\end{figure}
\begin{table}[H]
\caption{E5 benchmark (task-incremental, three seeds): ACC / BWT.}
\centering
\scriptsize
\begin{tabular}{lcccccc}
\toprule
 & Fine-tune & EWC & A-GEM & $\rho$-EWC & $\rho$-EWC+replay & DER++ \\
\midrule
Split-CIFAR-10 & 0.536 / $-$0.480 & 0.526 / $-$0.492 & 0.773 / $-$0.184 & 0.569 / $-$0.436 & 0.815 / $-$0.111 & 0.851 / $-$0.068 \\
Split-CIFAR-100 & 0.229 / $-$0.522 & 0.349 / $-$0.330 & 0.460 / $-$0.259 & 0.243 / $-$0.501 & 0.531 / $-$0.117 & 0.596 / $-$0.071 \\
\bottomrule
\end{tabular}
\end{table}
Gated tests (three seeds): $\Omega$-gated EWC-DR vs.\ EWC-DR $\Delta$ACC $+2.41$ pp $[-3.25,5.71]$, $\Delta$BWT $+1.16$ pp $[-6.60,5.07]$; gradient-gated replay vs.\ matched-mean replay $\Delta$ACC $-1.44$ pp $[-8.10,3.33]$. DER++ in the same run reaches ACC $0.862$. A custom class-incremental Seq-CIFAR-100 projection test moves ACC from $0.107$ to $0.121$ and BWT from $-0.856$ to $-0.841$ over three seeds at poor absolute accuracy; we report it only as a sign.

\subsection{Robustness}
\label{app:robust}
The robustness studies vary 1 part of the pipeline at a time and recompute the compatibility--forgetting correlation on the 20 Split-CIFAR-10 transitions. The estimator grid varies the number of probe examples and shuffled references and leaves the correlation between $-0.83$ and $-0.66$ in all 24 settings; generic representation similarity (CKA) and plain probe accuracy do much worse, indicating that the normalized, directional construction carries the relevant signal. The label study replaces the full labelled sample with a few labels per class; agreement with the full score rises quickly with the label budget, while label-free variants lose the signal. The optimizer and pretrained-backbone studies retrain the residents under different regimes and initializations.
\begin{figure}[H]
\centering
\includegraphics[width=\linewidth]{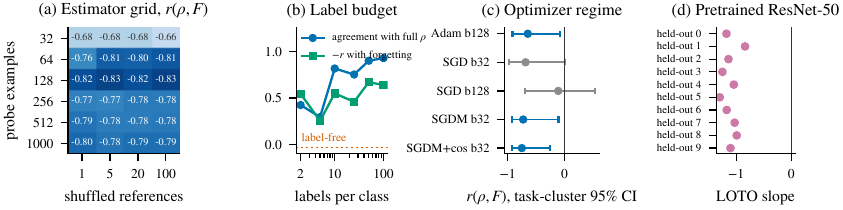}
\caption{Robustness. (a) All 24 estimator settings. (b) Label budget: agreement with the full score and correlation with forgetting; red dotted: label-free k-means variant. (c) Optimizer regimes with task-cluster intervals (grey: interval includes zero). (d) Leave-one-task-out slopes on a pretrained ResNet-50.}
\end{figure}
Optimizer regimes (Pearson $r$, task-cluster 95\% CI, negative LOTO slopes): Adam $-0.646$ $[-0.91,-0.08]$, 5/5; SGD b32 $-0.685$, 5/5; SGD b128 $-0.111$, 4/5; SGDM $-0.726$, 5/5; SGDM+cosine $-0.749$, 5/5. Label budgets (agreement with full score / $r$ with forgetting): 2: $0.42/-0.54$; 5: $0.29/-0.25$; 10: $0.82/-0.55$; 25: $0.75/-0.46$; 50: $0.90/-0.67$; 100: $0.93/-0.64$; label-free k-means $0.05/+0.03$; label-free CKA $-0.44/+0.03$. A single-head class-incremental construction gives $r=-0.201$ ($p=0.396$); the results are therefore stated for the task-incremental setting.

\subsection{Scale tests}
\label{app:scale}
The scale tests examine whether a literal feature-level ecological model can replace the transition-level coordinate. All run in a regime with substantial forgetting, so a failure cannot be attributed to a lack of signal. Jacobian-derived competition coefficients are tested at the task level and per feature, with probe sets of 64, 256, and 1024 examples at two layers. Lotka--Volterra dynamics are fitted with coefficients fixed from the Jacobians and, separately, within short windows where the coefficients might be locally stable, and both are compared with constant-rate and shuffled-coefficient baselines on held-out trajectory segments. Resource-competition ($R^\ast$) constructions and a niche/fitness decomposition are tested as coexistence mechanisms. None reaches its pass criterion. The transition-level coordinate orders forgetting on the same transitions, which motivates the mesoscopic interpretation in \S\ref{sec:e6}.
\begin{table}[H]
\caption{Pre-registered microscopic tests (E6). All run in a regime with forgetting s.d.\ $0.064$.}
\centering
\scriptsize
\begin{tabular}{p{2.0in}p{1.6in}p{1.4in}}
\toprule
Test & Criterion & Result \\
\midrule
Jacobian competition, task level & $r\ge0.5$, perm.\ $p\le0.05$ & $r=0.177$, $p=0.54$ \\
Jacobian competition, feature level & median partial $r>0.10$ & $\le0.105$ at 1024 probes; $\le6.7\%$ significant \\
Fixed-coefficient LV & beats baselines (Wilson $>0.5$) & 0/60; median held-out $R^2=-29.2$ \\
Locally stationary LV & stable windows exist & 0/330 stable; unstable $R^2=-1.08$ \\
Neural $R^\ast$ & resources identifiable & 1.9\%; precision 0.0 vs.\ base 0.58 \\
Niche/fitness (ND--FD) decomposition & explains forgetting & $R^2=0.21$ (held-out $0.11$), 45 pairs \\
\bottomrule
\end{tabular}
\end{table}

\end{document}